\documentclass[journal,onecolumn,oneside]{IEEEtran}

\usepackage[pdftex]{graphicx}
\DeclareGraphicsExtensions{.pdf,.jpeg,.jpg,.png}
\usepackage[caption = false,font = footnotesize]{subfig}
\usepackage{cite}
\usepackage{url}
\usepackage{rotating}
\usepackage[subpreambles=true]{standalone}

\usepackage[cmex10]{amsmath}
\usepackage{amsfonts}
\usepackage{amssymb}
\usepackage{amsthm}
\usepackage{bm}
\usepackage{empheq}
\usepackage{booktabs}

\usepackage{algorithmic}
\usepackage{setspace}
\usepackage{array}

\usepackage{url}
\usepackage[table]{xcolor}
\usepackage{multirow}
\usepackage{booktabs}
\usepackage{float}
\usepackage[flushleft]{threeparttable} 
\usepackage{balance}

\newsubfloat{figure}

\newcommand{\vect}[1]{\mathbf{#1}}

\DeclareMathSymbol{\R}{\mathalpha}{AMSb}{"52}

\begin{document}

\title{Complex-valued Neural Networks with \\ Non-parametric Activation Functions}

\author{Simone~Scardapane,~\IEEEmembership{Member,~IEEE,}
        Steven~Van~Vaerenbergh,~\IEEEmembership{Senior~Member,~IEEE,}
        Amir~Hussain,~\IEEEmembership{Senior~Member,~IEEE,}
        and~Aurelio~Uncini,~\IEEEmembership{Member,~IEEE}
\thanks{S. Scardapane and A. Uncini are with the Department of Information Engineering, Electronics and Telecommunications (DIET), Sapienza University of Rome, Via Eudossiana 18, 00184 Rome, Italy. Emails: {simone.scardapane,aurelio.uncini}@uniroma1.it}
\thanks{S. Van Vaerenbergh is with the Department of Communications Engineering, University of Cantabria, Av. los Castros s/n, 39005 Santander, Cantabria, Spain. Email: steven.vanvaerenbergh@unican.es.}%
\thanks{A. Hussain is with the Division of Computing Science \& Maths, School of Natural Sciences, University of Stirling, Stirling FK9 4LA, Scotland, UK.. Email: ahu@cs.stir.ac.uk.}%
}

\markboth{Preprint submitted to IEEE Trans. on Emerging Topics in Computational Intelligence}%
{Scardapane \MakeLowercase{\textit{et al.}}: Complex-valued NNs with non-parametric AFs}

\maketitle

\begin{abstract}
Complex-valued neural networks (CVNNs) are a powerful modeling tool for domains where data can be naturally interpreted in terms of complex numbers. However, several analytical properties of the complex domain (e.g., holomorphicity) make the design of CVNNs a more challenging task than their real counterpart. In this paper, we consider the problem of flexible activation functions (AFs) in the complex domain, i.e., AFs endowed with sufficient degrees of freedom to adapt their shape given the training data. While this problem has received considerable attention in the real case, a very limited literature exists for CVNNs, where most activation functions are generally developed in a split fashion (i.e., by considering the real and imaginary parts of the activation separately) or with simple phase-amplitude techniques. Leveraging over the recently proposed kernel activation functions (KAFs), and related advances in the design of complex-valued kernels, we propose the first fully complex, non-parametric activation function for CVNNs, which is based on a kernel expansion with a fixed dictionary that can be implemented efficiently on vectorized hardware. Several experiments on common use cases, including prediction and channel equalization, validate our proposal when compared to real-valued neural networks and CVNNs with fixed activation functions.
\end{abstract}

\begin{IEEEkeywords}
Neural networks, Activation functions, Kernel methods, Complex domain.
\end{IEEEkeywords}

\IEEEpeerreviewmaketitle

\section{Introduction}
\label{sec:introduction}

\IEEEPARstart{O}{ver} the last years, machine learning techniques have obtained impressive results in a wide range of fields, especially when dealing with supervised problems \cite{lecun2015deep,8264962,zheng2017video}. The majority of these applications has focused on the case of \textit{real-valued} data: as an example, most of the deep learning frameworks currently used today can only work with floating point (or integer) numbers. Several applicative domains of interest, however, exhibit data that can be more naturally modeled using \textit{complex-valued} algebra, from image processing to time-series prediction, bioinformatics, and robotics' control (see \cite{hirose2003complex,schreier2010statistical} for a variety of examples). While complex data can immediately be transformed to a real domain by considering the real and imaginary components separately, the resulting loss of phase information gives rise to algorithms that are generally less efficient (or expressive) than alternative methods able to work directly in the complex domain, as evidenced by a large body of literature \cite{mandic2007complex}. Due to this, many learning algorithms have been extended to deal with complex data, including linear adaptive filters \cite{fisher1983complex,schreier2010statistical}, kernel methods \cite{bouboulis2011extension,tobar2012novel,boloix2017widely}, component analysis \cite{scarpiniti2008generalized}, and neural networks (NNs) \cite{georgiou1992complex,kim2003approximation,arjovsky2016unitary,danihelka2016associative,guberman2016complex,trabelsi2017deep,8109745}. We consider this last class of algorithms in this paper.

Despite the apparent similarity between the real and complex domains, working directly in the latter is challenging because of several non-intuitive analytical properties of the complex algebra. Most notably, almost all cost functions involved in the training of complex models require non-analytic (also known as non-holomorphic \cite{bouboulis2011extension}) functions, so that standard complex derivatives cannot be used in the definition of the optimization algorithms. This is why several algorithms defined before the last decade considered optimizing the real and imaginary components separately, resulting in a more cumbersome notation which somehow hindered their development \cite{leung1991complex}. More recently, this problem has been solved by the adoption of the so-called CR-calculus (or Wirtinger's calculus), allowing to define proper complex derivatives even when dealing with non-analytic functions \cite{brandwood1983complex,kreutz2009complex}, by considering explicitly their dependence on both their arguments and their complex conjugates. We describe CR-calculus more in depth in Section \ref{sec:preliminaries}.

When dealing with neural networks, another challenging task concerns the design of a proper activation function in the complex domain. In the real-valued case, the use of the rectified linear unit (ReLU) has been instrumental in the development of truly deep networks \cite{glorot2011deep,maas2013rectifier}, and has spun a wave of further research in the topic, e.g., see \cite{klambauer2017self,ramachandran2017swish} for very recent examples. In the complex case, Liouville's theorem asserts that the only complex function which is analytic and bounded at the same time is a constant one. Due to the preference for bounded activation functions before the introduction of the ReLU, many authors in the past preferred bounded functions to analytic ones, most notably in a split organization, wherein the real and independent parts of the activations are processed separately \cite{nitta1997extension}, or in a phase-amplitude configuration, in which the nonlinearity is applied only to the magnitude component, while the phase component is preserved \cite{georgiou1992complex}. Even extending the ReLU function to the complex domain has been shown to be non-trivial, and several authors have proposed different variations \cite{guberman2016complex,arjovsky2016unitary}.

In this paper, we consider the problem of \textit{adapting} activation functions in the complex domain. For real-valued NNs, there is a large body of literature pointing to the fact that endowing activation functions with several degrees of freedom can improve the accuracy of the trained networks, ease the flow of the back-propagated gradient, or vastly simplify the design of the network. In the simplest case, we can consider parametric functions having only a few (generally less than three) parameters per neuron, such as the parametric ReLU \cite{he2015delving}, the S-shaped ReLU \cite{jin2016deep}, or the self-normalizing exponential linear unit (SELU) \cite{klambauer2017self}. More in general, we can think of \textit{non-parametric} activation functions, that can adapt to potentially any shape in a purely data-driven fashion, with a flexibility that can be controlled by the user, and to which standard regularization techniques can be applied. In the real-valued case, a lot of research has been devoted to the topic, including the design of Maxout networks \cite{goodfellow2013maxout}, adaptive piecewise linear (APL) units \cite{agostinelli2014learning}, spline functions \cite{scardapane2016learning}, and the recently proposed kernel activation functions (KAFs) \cite{scardapane2017kafnets}. When dealing with complex-valued NNs (CVNNs), however, only a handful of works have considered adapting the activation functions \cite{scarpiniti2008generalized,trabelsi2017deep}, and only in the simplified parametric case, or when working in a split configuration. In this sense, how to design activation functions that can adapt to the training data while remaining simple to implement remains an open question.

\subsection*{Contributions of the paper}
In this paper, we significantly extend KAFs \cite{scardapane2017kafnets} in order to design non-parametric activation functions for CVNNs. The basic idea of KAFs is to exploit a kernel expansion at every neuron, in which the elements of the kernel dictionary are fixed beforehand, while the mixing coefficients are adapted through standard optimization techniques. As described in \cite{scardapane2017kafnets}, this results in functions that are universal approximators, smooth over their entire domain, and whose implementation can leverage highly vectorized CPU/GPU libraries for matrix multiplication. 

Here, we propose two different techniques to apply the general idea of KAFs in the context of CVNNs. In the first case, we use a split combination where the real and the imaginary components are processed by two independent KAFs sharing the same dictionary. In the second case, we leverage recent works on complex-valued reproducing kernel Hilbert spaces \cite{bouboulis2011extension} to redefine the KAF directly in the complex domain, by describing several choices for the kernel function. We show via multiple experimental comparisons that CVNNs endowed with complex-valued KAFs can outperform both real-valued NNs and CVNNs having only fixed or parametric activation functions.

\subsection*{Organization of the paper}
In Section \ref{sec:preliminaries} we introduce the basic theoretical elements underpinning optimization in a complex domain and CVNNs. Then, in Section \ref{sec:complex_valued_afs} we summarize research on designing activation functions for CVNNs. The two proposed complex KAFs are given in Section \ref{sec:split_kaf} (split KAF) and Section \ref{sec:complex_kaf} (fully complex KAF). We provide an experimental evaluation in Section \ref{sec:experimental_evaluation} before concluding in Section \ref{sec:conclusions}.

\subsection*{Notation}
We denote vectors using boldface lowercase letters, e.g., $\vect{a}$; matrices are denoted by boldface uppercase letters, e.g., $\vect{A}$. All vectors are assumed to be column vectors. A complex number $z \in \mathbb{C}$ is represented as $z = a + ib$, where $a=\Re\{z\}$ and $b=\Im\{z\}$ are, respectively, the real part and the imaginary part of the number, and $i = \sqrt{-1}$. Sometimes, we also use $z_r$ and $z_i$ to denote the real and imaginary parts of $z$ for simplicity. Magnitude and phase of a complex number are given by $\lvert z \rvert$ and $\phi(z)$ respectively. $z^* = a - ib$ denotes the complex conjugate of $z$. Other notation is introduced in the text when appropriate.

\section{Preliminaries}
\label{sec:preliminaries}

\subsection{Complex algebra and CR-calculus}

We start by introducing the basic theoretical concepts required to define a complex-valued function and to optimize it. We consider scalar functions first, and discuss the multivariate extension later on. Any complex-valued function $f: \mathbb{C} \rightarrow \mathbb{C}$ can be written as:
\begin{equation}
f(z) = u(a,b) + iv(a,b) \,,
\end{equation}
where $u(\cdot, \cdot)$ and $v(\cdot, \cdot)$ are real-valued functions in two arguments. The function $f$ is said to be \textit{real-differentiable} if the partial derivatives of $u$ and $v$ with respect to $a$ and $b$ are defined. Additionally, the function is called \textit{analytic} (or holomorphic) if it satisfies the Cauchy-Riemann conditions:
\begin{equation}
\frac{\partial u(a,b)}{\partial a} = \frac{\partial v(a,b)}{\partial b} \;\; \text{and} \;\; \frac{\partial v(a,b)}{\partial a} = -\frac{\partial u(a,b)}{\partial b} \,.
\label{eq:cauchy_riemann_conditions}
\end{equation}
Only analytic functions admit a complex derivative in the standard sense, but most functions used in practice for CVNNs do not satisfy \eqref{eq:cauchy_riemann_conditions} (e.g., functions with real-valued outputs for which $v(a,b)=0$ everywhere). In this case, CR-calculus \cite{kreutz2009complex} provides a theoretical framework to handle non-analytic functions directly in the complex domain without the need to switch back and forth between definitions in the complex domain and gradients' computations in the real one.

The main idea is to consider $f$ explicitly as a function of both $z$ and its complex conjugate $z^* = a - ib$, which we denote as $f(z,z^*)$. If $f$ is real-differentiable, then it is also analytic with respect to $z$ when keeping $z^*$ constant and vice versa. Thus, we can define a pair of (complex) derivatives as follows \cite{brandwood1983complex,kreutz2009complex}:
\begin{align}
\text{R-derivative} \triangleq \left.\displaystyle\frac{\partial f(z,z^*)}{\partial z} \right\rvert_{z^*=\text{const}}  = \frac{1}{2}\left( \frac{\partial f}{\partial a} - i\frac{\partial f}{\partial b} \right) \,, \\
\text{R*-derivative} \triangleq \left.\displaystyle\frac{\partial f(z,z^*)}{\partial z^*} \right\rvert_{z=\text{const}} = \frac{1}{2}\left( \frac{\partial f}{\partial a} + i\frac{\partial f}{\partial b} \right) \,. 
\end{align}
Everything extends to multivariate functions $f: \mathbb{C}^n \rightarrow \mathbb{C}$ of a complex vector $\vect{z} \in \mathbb{C}^n$ by defining the cogradient and conjugate cogradient operators:
\begin{eqnarray}
\nabla_{\vect{z}} & = \displaystyle \left( \frac{\partial}{\partial z_1}, \ldots, \frac{\partial}{\partial z_n}\right)^T \,, \\
\nabla_{\vect{z}^*} & = \displaystyle \left( \frac{\partial}{\partial z^*_1}, \ldots, \frac{\partial}{\partial z^*_n}\right)^T \,.
\end{eqnarray}
Then, a necessary and sufficient condition for $\vect{z}_0$ to be a minimum of $f$ is either $\nabla_\vect{z_0} f(\vect{z}_0, \vect{z}_0^*) = 0$ or $\nabla_\vect{z^*_0} f(\vect{z}_0, \vect{z}_0^*) = 0$ \cite{brandwood1983complex}. CR-calculus inherits most of the standard properties of the real derivatives, including the chain rule and the differential rule, e.g., see \cite{kreutz2009complex}. For the important case where the output of the function is real-valued (as is the case for the loss function when optimizing CVNNs) we have the additional property:
\begin{equation}
\Bigl( \nabla_{\vect{z}} f(\vect{z}, \vect{z}^*) \Bigr)^* = \nabla_{\vect{z}^*} f(\vect{z}, \vect{z}^*) \,.
\end{equation}
Combined with the Taylor's expansion of the function, an immediate corollary of this property is that the direction of steepest ascent of $f$ in the point $\vect{z}$ is given by the \textit{conjugate} cogradient operator evaluated in that point \cite{kreutz2009complex}. Up to a multiplicative constant term, this result coincides with taking the steepest descent direction with respect to the real derivatives, allowing for a straightforward implementation in most optimization libraries.

\subsection{Complex-valued neural networks}

We now turn our attention to the approximation of multivariate complex-valued functions. A generic CVNN is composed by stacking $L$ layers via the alternation of linear and nonlinear operations. In particular, the $l$-th layer is described by the following equation:
\begin{equation}
\vect{h}_l = g \left( \vect{W}_l\vect{h}_{l-1} + \vect{b}_l \right) \,,
\label{eq:nn_layer}
\end{equation}
where $\vect{h}_{l-1} \in \mathbb{C}^{N_{l-1}}$ is the $N_{l-1}$-dimensional input to the layer, $\vect{W}_l \in \mathbb{C}^{N_{l-1} \times N_l}$ and $\vect{b}_l \in \mathbb{C}^{N_l}$ are adaptable weight matrices, and $g(\cdot)$ is a (complex-valued) activation function applied element-wise, which will be discussed more in depth later on. By definition, $\vect{x} = \vect{h}_0$ denotes the input to the network, while $\hat{y} = h_L$ denotes the final output, which we assume one-dimensional for simplicity. Some results on the approximation properties of this model are given in \cite{kim2003approximation}, while \cite{trabelsi2017deep} describes some techniques to initialize the adaptable linear weights in the complex domain.

Given $I$ input/output pairs $\mathcal{S} = \left\{ \vect{x}_n, y_n \right\}_{n=1}^I$, we train the CVNN by minimizing a cost function given by:
\begin{equation}
J(\vect{w}) = \sum_{n=1}^I l(y_n, \hat{y}_n) \,,
\label{eq:global_cost_function}
\end{equation}
where $\vect{w} \in \mathbb{C}^Q$ collects all the adaptable weights of the network and $l(\cdot, \cdot)$ is a loss function, e.g., the squared loss:
\begin{equation}
l(y, \hat{y}) = \lvert y - \hat{y} \rvert^2 = \left(y - \hat{y}\right)\left(y - \hat{y}\right)^* \,.
\label{eq:squared_loss}
\end{equation}
Following the results described in the previous section, a basic steepest descent approach to optimize \eqref{eq:global_cost_function} is given by the following update equation at the $t$-th iteration:
\begin{equation}
\vect{w}_{t+1} = \vect{w}_{t} - \mu \nabla_{\vect{w}^*} J(\vect{w}, \vect{w}^*) \,, 
\end{equation}
where $\mu \in \R$ is the learning rate. More in general, we can use noisy versions of the gradient given by sampling a mini-batch of elements, or accelerate the optimization process by adapting most of the state-of-the-art techniques used for real-valued neural networks \cite{bottou2016optimization}. We can also apply some techniques that are specific to the complex domain. For example \cite{xu2015convergence}, inspired by the theory of widely linear adaptive filters, augments the input to the CVNN with its complex conjugate $\vect{x}^*$. Additional improvements can be obtained by replacing the real-valued $\mu$ with a complex-valued learning rate \cite{zhang2016complex}, which can speed up convergence in some scenarios.

\section{Complex-valued activation functions}
\label{sec:complex_valued_afs}

\begin{figure*}
\subfloat[Absolute value]{
\includegraphics[width=0.48\columnwidth,keepaspectratio]{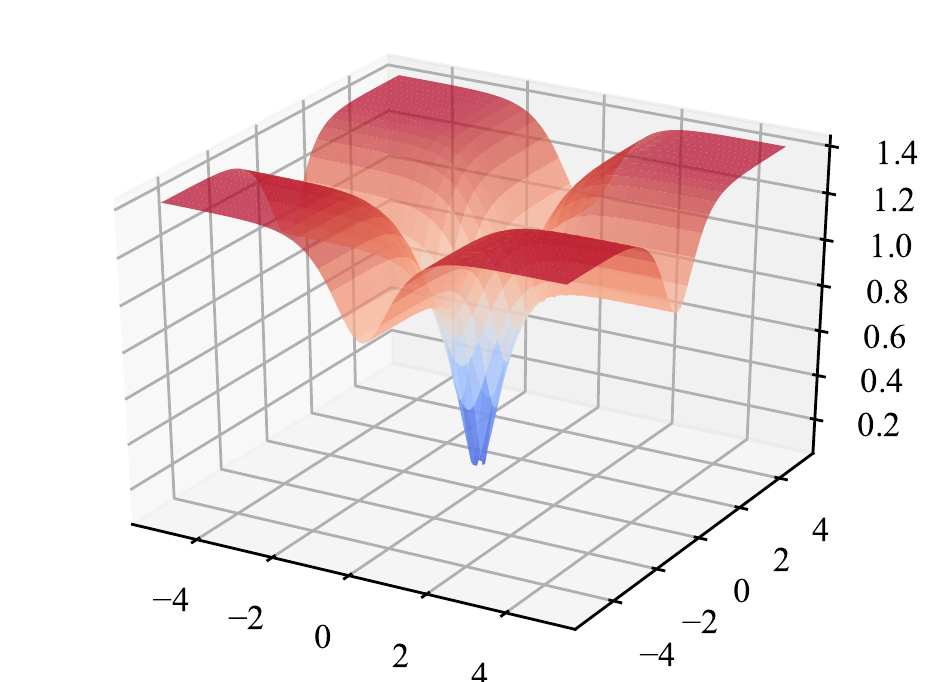}
\label{fig:split_tanh_magnitude}
} \hfil
\subfloat[Phase]{
\includegraphics[width=0.48\columnwidth,keepaspectratio]{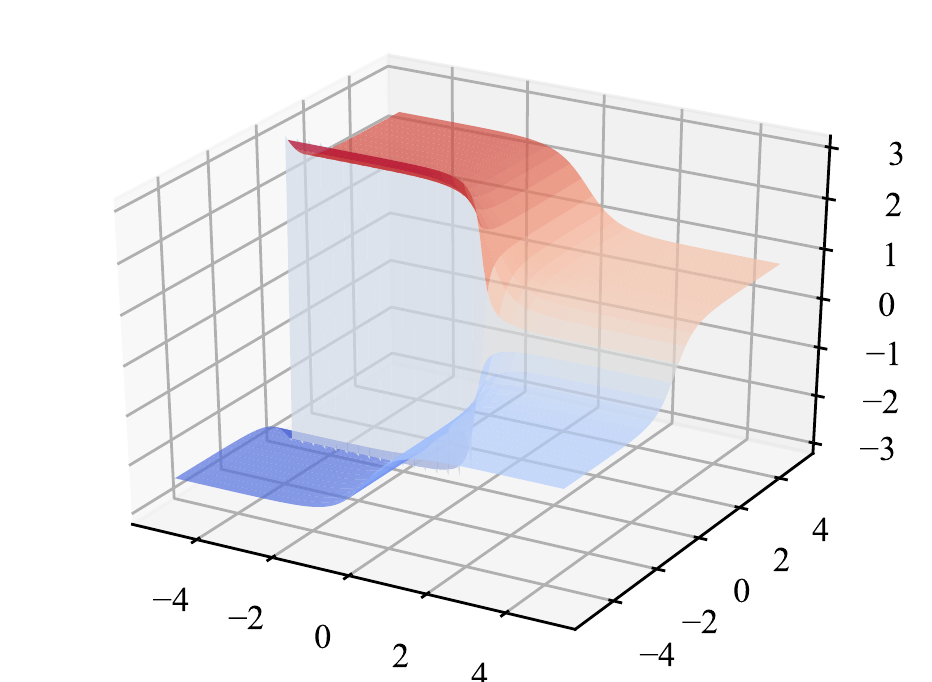}
\label{fig:split_tanh_phase}
}
\caption{Example of split activation function having $g_R(\cdot) = \tanh(\cdot)$ in \eqref{eq:split_activation_function} processing both the real and the imaginary parts of the input. (a) Magnitude of the output. (b) Phase of the output.}
\label{fig:split_tanh}
\end{figure*}

As we stated in the introduction, choosing a proper activation function in \eqref{eq:nn_layer} is more challenging than in the real case because of Liouville's theorem, stating that the only complex-valued functions that are bounded and analytic everywhere are constants. So in practice, one need to choose between boundedness and analyticity. Before the introduction of the ReLU activation \cite{glorot2011deep}, almost all activation functions in the real case where bounded. Consequently, initial approaches to design CVNNs always preferred non-analytic functions in order to preserve boundedness, most commonly by applying real-valued activation functions separately to the real and imaginary parts \cite{nitta1997extension}:
\begin{equation}
g(z) = g_R(\Re\left\{z\right\}) + i g_R(\Im\left\{z\right\}) \,,
\label{eq:split_activation_function}
\end{equation}
where $z$ is a generic input to the activation function in \eqref{eq:nn_layer}, and $g_R(\cdot)$ is some real-valued activation function, e.g., sigmoid. This is called a \textit{split activation function}. As a representative example, the magnitude and phase of the split-$\tanh$ when varying the activation are given in Fig. \ref{fig:split_tanh}. Early proponents of this approach can be found in \cite{benvenuto1992complex} and \cite{leung1991complex}.

Another common class of non-analytic activation functions are the phase-amplitude (PA) functions popularized by \cite{georgiou1992complex,hirose1992continuous}:

\begin{align}
g(z) & = \frac{z}{c + \lvert z \rvert / r} \,, \label{eq:amp_activation_function} \\
g(z) & = \tanh\left\{ \frac{\lvert z \rvert}{m}\right\}\exp\left\{ i\phi(z) \right\} \,,
\end{align}
where $\phi(z)$ is the phase of $z$, while $c$, $r$ and $m$ are positive constant which in most cases are set equal to $1$. PA functions can be seen as the natural generalization of real-valued squashing functions such as the sigmoid, because the output $g(z)$ has bounded magnitude but preserves the phase of $z$. 

A third alternative is to use fully-complex activation functions that are analytic and bounded almost everywhere, at the cost of introducing a set of singular points. Among all possible transcendental functions, it is common to consider the complex-valued extension of the hyperbolic tangent, defined as \cite{kim2003approximation}:
\begin{equation}
g(z) = \tanh\left\{z\right\} = \frac{\exp\left\{z\right\} - \exp\left\{-z\right\}}{\exp\left\{z\right\} + \exp\left\{-z\right\}} \,,
\end{equation}
possessing periodic singular points at the imaginary points $i\left(0.5 + n\right)\pi$, with $n \in \mathbb{N}$. However, careful scaling of the inputs and of the initial weights allows to avoid these singularities during training.

Finally, several authors have proposed extensions of the real-valued ReLU function $\text{ReLU}(s) = \max\left\{0, s\right\}$, motivated by the fact that its success in the deep learning literature does not immediately translate to the complex-valued case, where using it in a split function as in \eqref{eq:split_activation_function} results in poor performance \cite{trabelsi2017deep}. \cite{guberman2016complex} propose a complex-valued ReLU as:
\begin{equation}
g(z) = 
	\begin{cases}
	z & \text{ if } \Re\left\{z\right\}, \Im\left\{z\right\} \ge 0 \,, \\
	0 & \text{ otherwise }
	\end{cases} \,.
\label{eq:complex_relu}
\end{equation}
Alternatively, inspired by the PA functions to maintain the phase of the activation value, \cite{arjovsky2016unitary} propose the following modReLU function:
\begin{equation}
g(z) = \text{ReLU}\left(\lvert z \rvert + b\right) \exp\left\{i \phi(z) \right\} \,,
\label{eq:modrelu}
\end{equation}
where $b$ is an adaptable parameter defining a radius along which the output of the function is $0$. Another extension, the complex cardioid, is advanced in \cite{virtue2017better}:
\begin{equation}
g(z) = \frac{1}{2}\Bigl(1 + \cos\left\{ \phi(z) \right\} \Bigr)z \,,
\label{eq:complex_cardioid}
\end{equation}
maintaining phase information while attenuating the magnitude based on the phase itself. For real-valued inputs, \eqref{eq:complex_cardioid} reduces to the ReLU.

Note that in all cases these proposed activation functions are fixed or endowed with a very small degree of flexibility (as in \eqref{eq:modrelu}). In the following sections we describe a principled technique to design non-parametric activation functions for use in CVNNs.

\section{Split kernel activation functions}
\label{sec:split_kaf}

Our first proposal is a split function as in \eqref{eq:split_activation_function}, where we use non-parametric (real-valued) functions for $g_R(\cdot)$ in place of fixed ones. Specifically, we consider the kernel activation function (KAF) proposed in \cite{scardapane2017kafnets}, which will also serve as a base for the fully complex-valued proposal of the following section. Here, we introduce the basic elements of the KAF, and we refer to the original paper \cite{scardapane2017kafnets} for a fuller exposition.

The basic idea of a KAF is to model each activation function as a one-dimensional kernel model, where the kernel elements are chosen in a proper way to obtain an efficient backpropagation step. Consider the generic activation function $g_R(s)$, where $s$ denotes either the real or the imaginary part of $z$ as in \eqref{eq:split_activation_function}. To obtain a flexible shape, we can model a linear predictor on a high-dimensional feature space $\Phi(s)$ of the activation. However, this process becomes infeasible for a large number of feature transformations, and cannot handle infinite-dimensional feature spaces. However, for feature maps associated to a reproducing kernel Hilbert space $\mathcal{H}$ with kernel $\kappa(\cdot,\cdot)$,\footnote{Remember that a function $\kappa(\cdot, \cdot)$ is a valid kernel function if it respects the positive semi-definiteness property, i.e., for any possible choice of $\left\{\alpha_n\right\}_{n=1}^D$ and $\left\{d_n\right\}_{n=1}^D$ in \eqref{eq:proposed_kaf} we have that:
\begin{equation}
\sum_{n=1}^D \sum_{m=1}^D \alpha_n \alpha_m \kappa\left(d_n, d_m\right) \ge 0 \,.
\label{eq:psd_kernel}
\end{equation}} we can write an equivalent linear model by exploiting the representer's theorem as:
\begin{equation}
g_R(s) = \sum_{n=1}^D \alpha_n \kappa\left(s, d_n\right) \,,
\label{eq:proposed_kaf}
\end{equation}
where $\left\{\alpha_n\right\}_{n=1}^D$ are the mixing coefficients and $\left\{d_n\right\}_{n=1}^D$ make up the so-called dictionary of the kernel expansion \cite{hofmann2008kernel,liu2011kernel}. In the context of a neural network, the dictionary elements cannot be selected \textit{a priori} because they would change at every step of the optimization algorithm depending on the distribution of the activation values. Instead, we exploit the fact that we are working with one-dimensional kernels to fix the elements beforehand, and we only adapt the mixing coefficients in the optimization step. In particular, we select the elements $d_1, \ldots, d_D$ by sampling $D$ values over the $x$-axis, uniformly around zero. In this way, the value $D$ becomes a hyper-parameter controlling the flexibility of the approach: for larger $D$ we obtain a more flexible method at the cost of a larger number of adaptable parameters. In general, since the function is only a small component of a much larger neural network, we have found values in $\left[10,20\right]$ to be sufficient for most applications. As the number of parameters per neuron can potentially grow without bound depending on the choice of $D$, we refer to such activation functions as non-parametric.

We use the same dictionary across the entire neural network, but two different sets of mixing coefficients for the real and imaginary parts of each neuron. Due to this, an efficient implementation of the proposed split-KAF is straightforward. In particular, consider the vector $\vect{z}$ containing the $N_l$ (complex) activations of a layer following the linear operations in \eqref{eq:nn_layer}. We build the matrix $\vect{K}_R \in \R^{N_l \times D}$ by computing all the kernel values between the real part of the activations and the elements of the dictionary (and similarly for $\vect{K}_I$ using the imaginary parts), and we compute the final output of the layer as:
\begin{equation}
\vect{h}_l = \left(\vect{A}_R \odot \vect{K}\right)\vect{1} + i \left(\vect{A}_I \odot \vect{K}_I\right)\vect{1} \,,
\label{eq:kaf_implementation}
\end{equation}
where $\odot$ represents element-wise product (Hadamard product), $\vect{A}_R, \vect{A}_I \in \R^{N_l \times D}$ are matrices collecting row-wise all the mixing coefficients for the real and imaginary components of the layer, and $\vect{1} \in \R^D$ is a vector of ones. If we need to handle batches of elements (or convolutive layers), we only need to slightly modify \eqref{eq:kaf_implementation} by adding additional trailing dimensions.

For all our experiments, we consider the 1D Gaussian kernel defined as:
\begin{equation}
\kappa(s, d_n) = \exp\left\{-\gamma\left(s - d_n\right)^2\right\} \,,
\label{eq:gaussian_kernel}
\end{equation}
where $\gamma \in \R$ is called the kernel bandwidth. In the proposed KAF scheme, the values of the dictionary are chosen according to a grid, and as such the optimal bandwidth parameter depends uniquely on the grid resolution. In particular, we use the following rule-of-thumb proposed in \cite{scardapane2017kafnets}:
\begin{equation}
\gamma = \frac{1}{6\Delta^2} \,,
\label{eq:sigma_rule_of_thumb}
\end{equation}
where $\Delta$ is the distance between the grid points. In order to provide an additional degree of freedom to our method, we also optimize a single $\gamma$ per layer via back-propagation after initializing it following \eqref{eq:sigma_rule_of_thumb}. 

\section{Fully-complex kernel activation functions}
\label{sec:complex_kaf}

While most of the literature on kernel methods in machine learning has focused on the real-valued case, it is well-known that the original mathematical treatment originated in the complex-valued domain \cite{aronszajn1950theory}. In the context of the kernel filtering literature, techniques to build complex-valued algorithms by separating the real and the imaginary components (as in the previous section) are called complexification methods \cite{bouboulis2011extension}. However, recently several authors have advocated for the direct use of (pure) complex-valued kernels leveraging the complex-valued treatment of RKHSs for a variety of fields, as surveyed in the introduction.

\begin{figure}
  \centering
  \includestandalone[mode=image, scale=1.2]{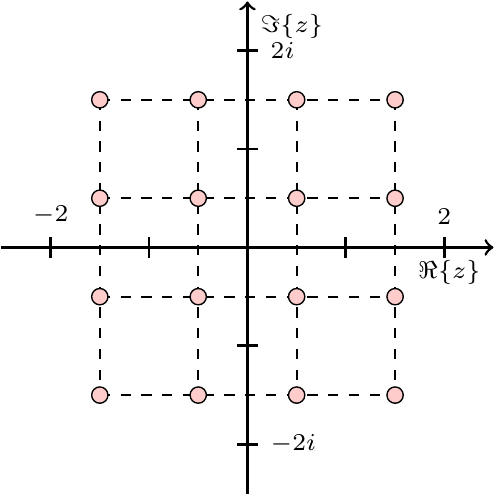}
  {\caption{A visual example of sampling the dictionary for the complex-valued KAF, in the complex plane, for $D=4$ in the range $\left[-1.5, 1.5\right]$.}
  \label{fig:complex_dictionary_sampling}}
\end{figure}

From a theoretical standpoint, defining complex RKHSs and kernels is relatively straightforward. As an example, a one-dimensional complex-function $\kappa_{\mathbb{C}}: \mathbb{C} \times \mathbb{C} \rightarrow \mathbb{C}$ is positive semi-definite if and only if:
\begin{equation}
\sum_{n=1}^D \sum_{m=1}^D \alpha_n^* \alpha_m \kappa\left(d_n, d_m\right) \ge 0 \,, \forall \alpha_n, \alpha_m, d_n, d_m \in \mathbb{C} \,,
\label{eq:psd_complex_kernel}
\end{equation}
where all values are now defined in the complex-domain. Any PSD function is then a valid kernel function. Based on this, in this paper we also propose a fully-complex, non-parametric KAF by defining \eqref{eq:proposed_kaf} directly in the complex domain, without the need for split functions:
\begin{equation}
g(z) = \sum_{n=1}^D \sum_{m=1}^D \alpha_{n,m} \kappa_{\mathbb{C}}\left(z, d_n + i d_m\right) \,,
\label{eq:proposed_complex_kaf}
\end{equation}
where the mixing coefficients $\left\{\alpha_{n,m}\right\}_{n,m=1}^D$ are now defined as complex numbers. Note that, in order for the dictionary to provide a dense sampling of the space of complex numbers, we now consider $D^2$ fixed elements arranged over a regular grid, an example of which is depicted in Fig. \ref{fig:complex_dictionary_sampling}. Due to this, we now have $D^2$ adaptable mixing coefficients per neuron, as opposed to $2D$ in the split case. We counter-balance this by selecting a drastically smaller $D$ (see the experimental section).

\begin{figure*}
\subfloat[Absolute value]{
\includegraphics[width=0.48\columnwidth,keepaspectratio]{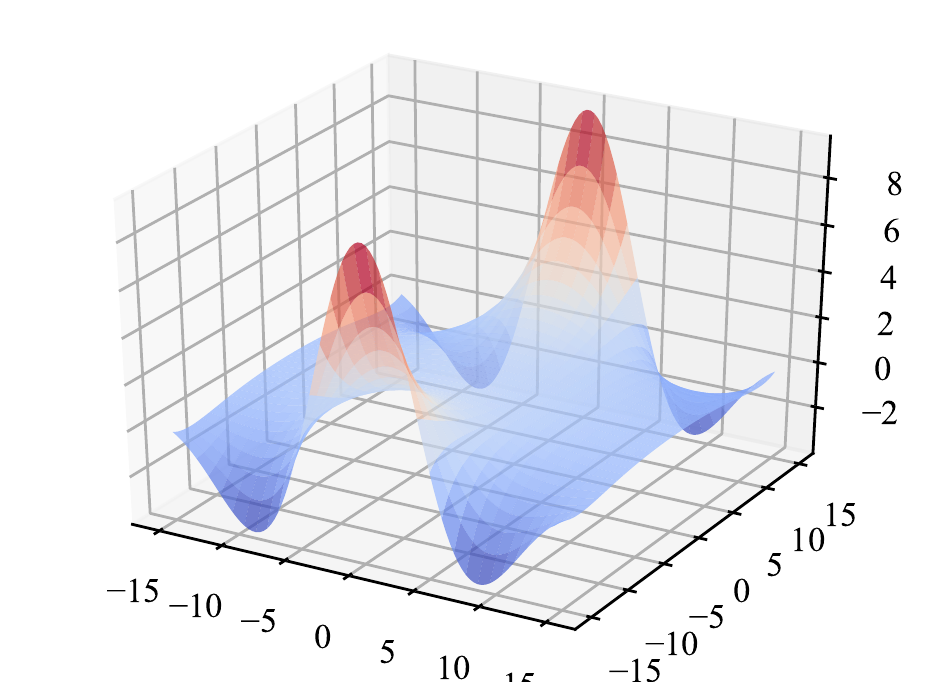}
\label{fig:complex_gaussian_kernel_real_part}
} \hfil
\subfloat[Phase]{
\includegraphics[width=0.48\columnwidth,keepaspectratio]{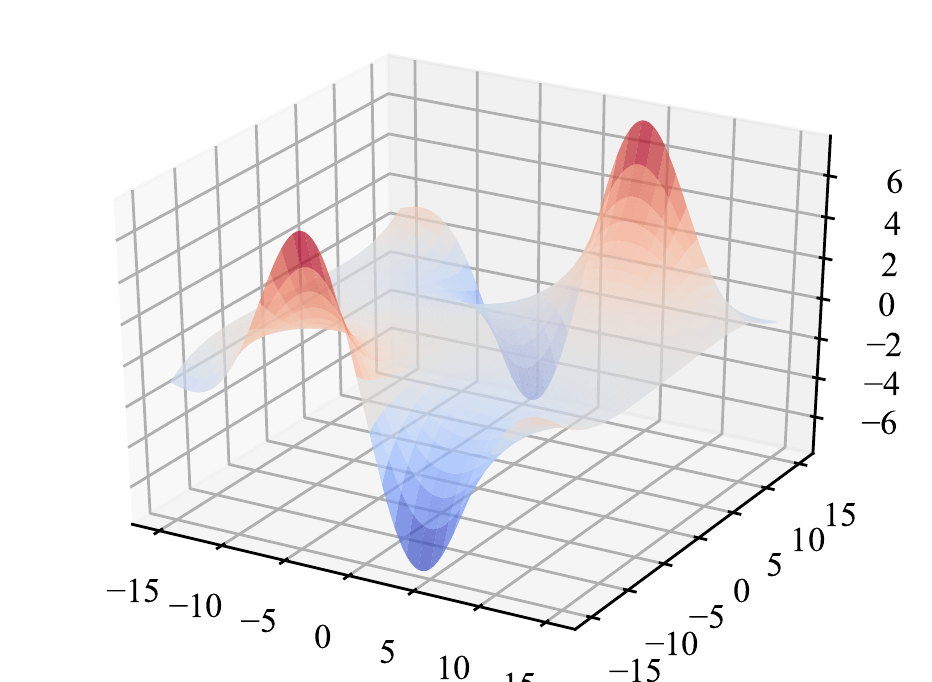}
\label{fig:complex_gaussian_kernel_imaginary_part}
}
\caption{Example of Gaussian complex kernel in \eqref{eq:complex_gaussian_kernel} with $d=0+i0$ and $\gamma=0.01$. Notice the scale of the axes (more details are provided in the text). (a) Real part of the output. (b) Imaginary part of the output.}
\label{fig:complex_gaussian_kernel}
\end{figure*}

An immediate complex-valued extension of the Gaussian kernel in \eqref{eq:gaussian_kernel} is given by:
\begin{equation}
\kappa_{\mathbb{C}}(z, d) = \exp\left\{-\gamma\left(z - d^* \right)^2\right\} \,,
\label{eq:complex_gaussian_kernel}
\end{equation}
where in our experiments the bandwidth hyper-parameter $\gamma$ is selected using the same rule-of-thumb as before and then adapted layer-wise. A complete analysis of the feature space associated to \eqref{eq:complex_gaussian_kernel} is given in \cite{steinwart2006explicit}. In order to gain some informal understanding, we can write the kernel explicitly in terms of the real and imaginary components of its arguments:
\begin{align}
\displaystyle \kappa_{\mathbb{C}}(z, d) & = \exp\left\{ - \gamma \lvert z_r  - d_r \rvert^2 \right\} \exp \left\{ \gamma \lvert z_i + d_i \rvert^2 \right\}  \nonumber \\
& \cdot \Bigl( \cos\left\{ 2\gamma\left( z_r - d_r \right)\left( z_i + d_i \right) \right\} \Bigr. \nonumber \\
& \Bigl. - i \sin\left\{ 2\gamma\left( z_r - d_r \right)\left( z_i + d_i \right) \right\} \Bigr) \,.
\end{align}
By analyzing the previous expression, we see that the complex-valued Gaussian kernel has several properties which are counter-intuitive if one is used to work with its real-valued restriction. First of all, \eqref{eq:complex_gaussian_kernel} cannot be interpreted as a standard similarity measure, because it depends on its arguments only via $\left( z_r - d_r \right)$ and $\left( z_i + d_i \right)$. For the same reasons, the kernel is not stationary, and it has an additional oscillatory behavior. We refer to Fig. \ref{fig:complex_gaussian_kernel} (or to \cite[Section IV-A]{boloix2017widely}) for an illustration of the kernel when fixing the second argument.

For these reasons, another extension of the Gaussian kernel to the complex domain is given in \cite{bouboulis2011extension}, where the authors propose to build a whole family of complex-valued kernels starting from any real-valued one $\kappa_{\mathbb{R}}$ as follows:
\begin{align}
\kappa_{\mathbb{C}}\left(z,d\right) & = \kappa_{\mathbb{R}}\left(z_r, d_r\right) + \kappa_{\mathbb{R}}\left(z_i, d_i\right) \nonumber \\
& + i\left( \kappa_{\mathbb{R}}\left(z_r, d_i\right) - \kappa_{\mathbb{R}}\left(z_i, d_r\right)\right) \,.
\label{eq:independent_kernel}
\end{align}
The new complex-valued kernel is called an \textit{independent} kernel. By plugging the real-valued Gaussian kernel \eqref{eq:gaussian_kernel} in the previous expression, we obtain a complex-valued expression that can still be interpreted as a similarity measure between the two points.

Note that several alternative kernels are also possible, many of which are specific to the complex-valued case, a prominent example being the Szego kernel \cite{bouboulis2011extension}:
\begin{equation}
\kappa_{\mathbb{C}}(z, d) = \frac{1}{\left(1 - zd^*\right)^2} \,.
\label{eq:szego_kernel}
\end{equation}

\section{Experimental evaluation}
\label{sec:experimental_evaluation}

In this section, we experimentally evaluate the proposed activation functions on several benchmark problems, including channel identification in Section \ref{sec:experiment_channel_identification}, wind prediction in Section \ref{sec:experiment_wind_prediction}, and multi-class classification in the complex domain in Section \ref{sec:experiment_digits_classification}. In all cases, we linearly preprocess the real and the imaginary components of the input features to lie in the $\left[-1,+1\right]$ range. We regularize all parameters with respect to their squared absolute value (which is equivalent to standard $\ell_2$ regularization applied on the real and imaginary components separately), but we exclude the bias terms and the window parameter in \eqref{eq:modrelu}. We select the strength of the regularization term and the size of the networks based on previous literature or on a cross-validation procedure, as described below. For optimization, we use a simple complex-valued extension of the Adagrad algorithm, which computes a per-parameter learning rate weighted by the squared magnitude of the gradients themselves. For each iteration, we construct a mini-batch by randomly sampling $40$ elements from the entire training dataset. All algorithms have been implemented in Python using the Autograd library \cite{maclaurin2015autograd}.

\subsection{Experiment 1 - Channel Identification}
\label{sec:experiment_channel_identification}

\begin{figure*}
\subfloat[Circular signal ($\rho=\frac{\sqrt{2}}{2}$)]{
\includegraphics[width=0.49\columnwidth,keepaspectratio]{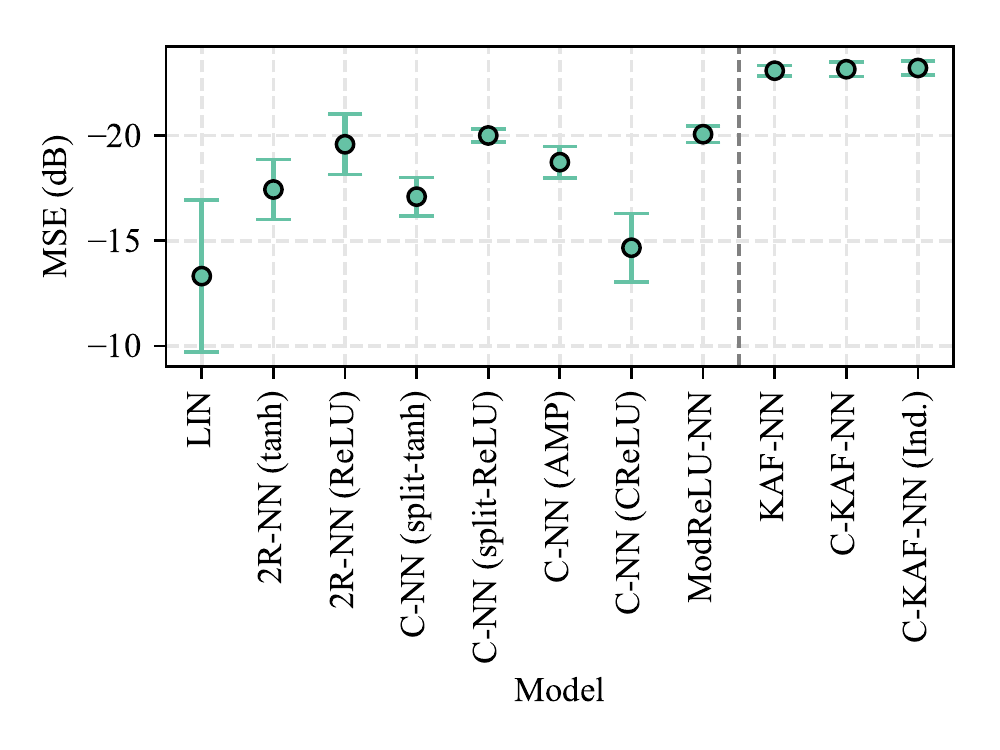}
\label{fig:results_experiment_1_circular}
} \hfil
\subfloat[Non-circular signal ($\rho=0.95$)]{
\includegraphics[width=0.49\columnwidth,keepaspectratio]{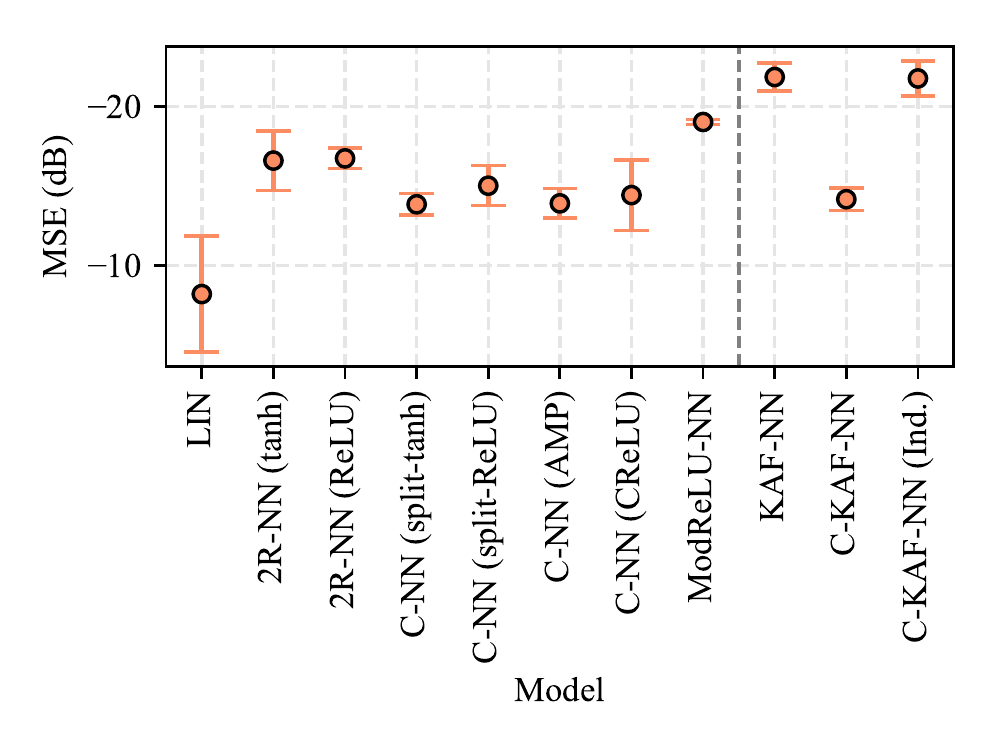}
\label{fig:results_experiment_1_noncircular}
}
\caption{Results for the first experiment, expressed in terms of MSE (dB). (a) Circular input signal. (b) Non-circular input signal. With a dashed line we divide the results of the proposed models.}
\label{fig:results_experiment_1}
\end{figure*}

Our first experiment is a standard benchmark in the complex-valued literature, i.e. a channel identification task \cite{bouboulis2015complex}. The input to the channel is generated as:
\begin{equation}
s_n = \left( \sqrt{1-\rho^2} X_n + i \rho Y_n \right) \,,
\end{equation}
where $X_n$ and $Y_n$ are Gaussian random variables, and the parameter $\rho$ determines the circularity\footnote{A random variable $Z$ is circular if $Z$ and $Z\exp\left\{i\psi\right\}$ have the same probability distribution for any angle $\psi$. Roughly speaking, non-circular signals are harder to predict, requiring the use of widely linear techniques when using standard linear filters \cite{bouboulis2011extension}.} of the signal. For $\rho=\frac{{\sqrt{2}}}{2}$ the input is circular, while for $\rho$ approaching $0$ or $1$ the signal is highly non-circular. The output of the channel is computed by first applying a linear filtering operation:
\begin{equation}
t_n = \sum_{k=1}^5 h(k) s_{n-k+1} \,,
\end{equation}
where:
\begin{align}
h(k) & = 0.432 \left( 1 + \cos \left\{\frac{2\pi(k-3)}{5}\right\} \right. \nonumber \\
& \left. - i\left( 1 + \cos\left\{ \frac{2\pi(k-3)}{10}\right\} \right) \right) \,,
\end{align}
for $k=1,\ldots,5$. Then, the output of the linear filter goes through a memoryless nonlinearity:
\begin{equation}
r_n = t_n + \left(0.15-i0.1\right)t_n^2 \,,
\end{equation}
and finally it is corrupted by adding white Gaussian noise in order to get the final signal $\tilde{r}_n$. The variance of the noise is selected to obtain a signal-to-noise ratio (SNR) of about $13$ dB. The input to the neural network is an embedding of channel inputs:
\begin{equation}
\vect{x} = \left[ s_{n-L+1}, s_{n-L+2}, \ldots, s_n \right]^T \,,
\end{equation}
with $L=5$, and the network is trained to output $\tilde{r}_n$. We generate $2000$ samples of the channel, and we randomly keep $15\%$ for testing, averaging over $15$ different generations of the dataset. We compare the following algorithms:
\begin{itemize}
\item \textbf{LIN}: a standard linear filter \cite{schreier2010statistical} with complex-valued coefficients.
\item \textbf{2R-NN}: a real-valued neural network taking as input the real and imaginary parts separately. For the activation functions in the hidden layers, we consider either a standard $\tanh$ or ReLUs.
\item \textbf{C-NN}: complex-valued neural networks with fixed activation functions, including a split-$\tanh$, a split-ReLU, the AMP function in \eqref{eq:amp_activation_function}, or the complex ReLU in \eqref{eq:complex_relu}.
\item \textbf{ModReLU-NN}: CVNN with adaptable activation functions with ModReLU neurons as in \eqref{eq:modrelu}. In this case, the coefficients of the neurons are all initialized at $0.1$ and later adapted.
\item \textbf{Proposed KAF-NN}: CVNN with the split-KAF proposed in Section \ref{sec:split_kaf}. We empirically select $D=20$ elements in the dictionary sampled uniformly in $\left[-2, +2\right]$.
\item \textbf{Proposed C-KAF-NN}: CVNN with the fully complex KAF proposed in Section \ref{sec:complex_kaf}. In this case, we test either the complex Gaussian kernel \eqref{eq:complex_gaussian_kernel}, or the independent kernel with the real Gaussian kernel as base. We empirically select $D=8$.
\end{itemize}
All algorithms are trained by minimizing the mean-squared error in \eqref{eq:squared_loss} on random mini-batches of $40$ elements. Following \cite{xu2015convergence}, in this scenario we consider one hidden layer with $10$ neurons (as more layers are not found to provide significant improvements in performance). The size of the regularization factor is empirically selected as $10^{-4}$. Results in terms of mean squared error (MSE) expressed in dBs are given in Table \ref{fig:results_experiment_1}, by considering either $\rho=\frac{\sqrt{2}}{2}$ (circular input signal) or the more challenging scenario $\rho=0.95$ (non-circular signal).

As expected, results are generally lower for the non-circular case, proportionally so for techniques that are not able to exploit the geometry of non-circular complex signals, such as non-widely linear models and real-valued neural networks. However, the proposed KAF-NN and C-KAF-NN are able to consistently out-perform all other methods in both scenarios in a stable fashion. Note that this difference in performance cannot be overcome by increasing the size of the other networks, thus pointing to the importance of adapting the activation functions also in the complex case. Interestingly, the complex Gaussian kernel in \eqref{eq:complex_gaussian_kernel} results in a poor performance, which is solved by using the independent one.

\subsection{Experiment 2 - Wind prediction}
\label{sec:experiment_wind_prediction}

\begin{figure*}
\centering
\subfloat[Absolute value]{
\includegraphics[width=0.45\columnwidth,keepaspectratio]{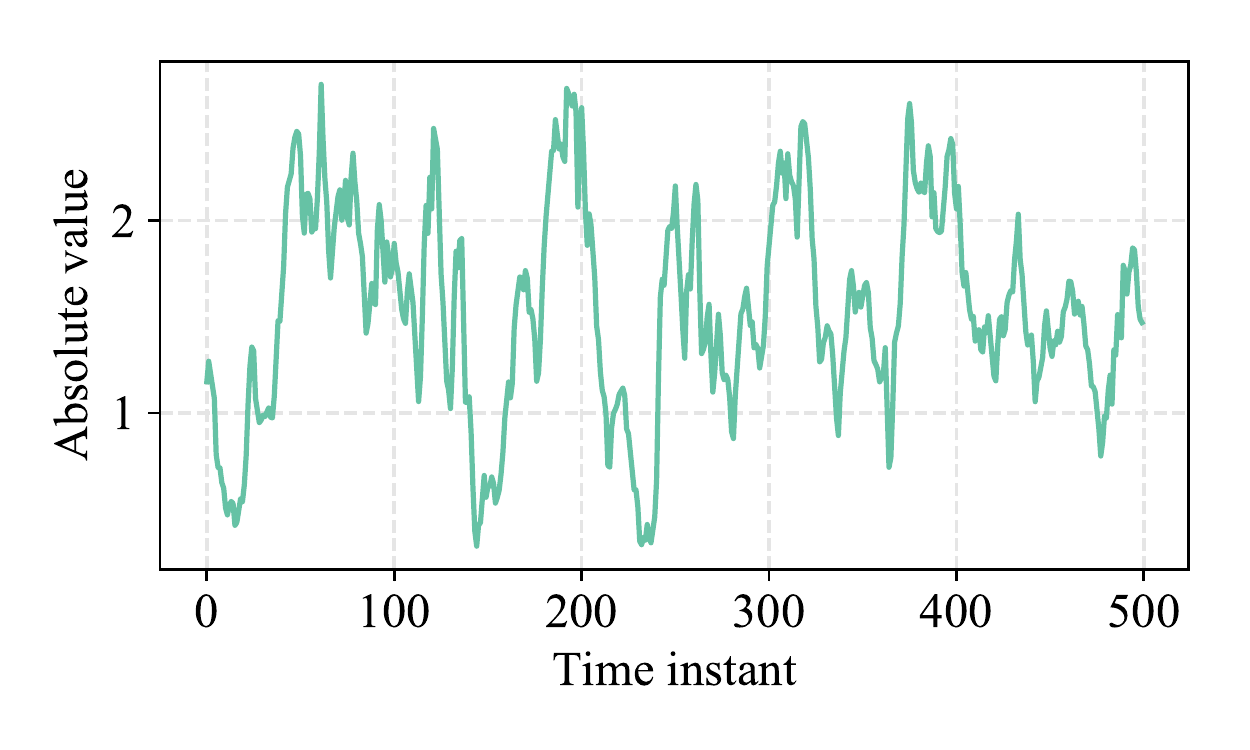}
\label{fig:wind_absolute_value}
} \hfil
\subfloat[Phase]{
\includegraphics[width=0.45\columnwidth,keepaspectratio]{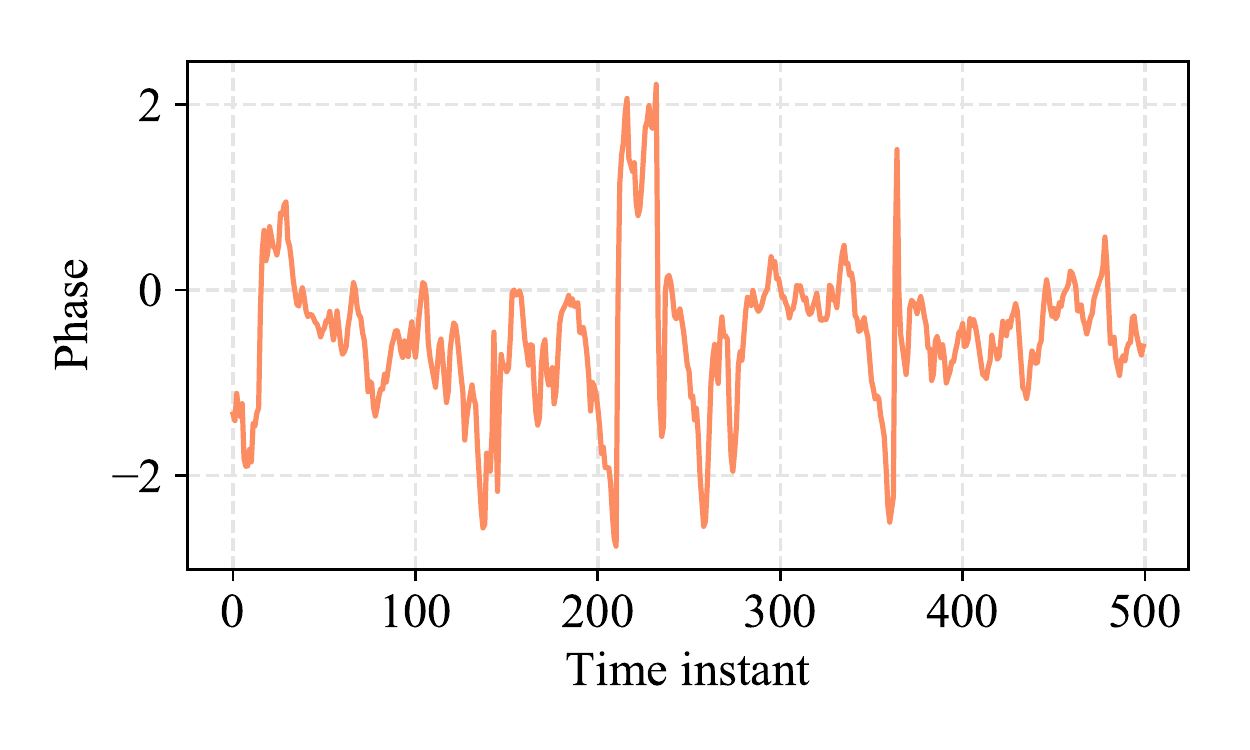}
\label{fig:wind_phase}
}
\caption{A plot of the complex-valued wind profile for the initial $500$ samples of the wind time-series. (a) Absolute value of the signal. (b) Phase of the signal.}
\label{fig:wind}
\end{figure*}

For the second experiment, we consider a real-world dataset for a task of wind prediction \cite{goh2006complex}. The dataset consists of $5000$ hourly samples of wind intensity collected along two different axes (north axis and east axis). The dataset is provided in three settings of wind regime, namely `low', `medium', and `high', from which we select the highest, being the most challenging one. In order to construct a complex-valued signal, the two samples for each hour are considered as the real and the imaginary components of a single complex number (for more motivation on the use of complex-valued information when dealing with wind forecasting, see \cite{goh2004complex,goh2005nonlinear,8109745,goh2006complex,kuh2009applications}. A snapshot of the absolute value and phase of the resulting signal is shown in Fig. \ref{fig:wind} for the initial $500$ samples. We consider the task of predicting both components of the wind for an $8$-hour-ahead horizon, starting from an embedding of the last $10$ hours of measurements. We select neural networks with $2$ hidden layers (as more hidden layers are not found to provide gain in performance), and we optimize both the number of neurons and the regularization factor on a held-out validation set. We test the datasets on the last $500$ components of the time-series, in terms of the $R^2$ coefficient of determination:

\begin{equation}
R^2 = 1 - \frac{\displaystyle\sum_{n=1}^{500} \lvert y_n - \hat{y}_n \rvert^2}{\displaystyle\sum_{n=1}^{500}\lvert y_n - \bar{y} \rvert^2} \,,
\end{equation}

\noindent where $y_n$ is the true value, $\hat{y}_n$ is the predicted value, and $\bar{y}$ is the mean of the true values computed from the test set. Positive values of $R^2$ denotes a prediction which is better than chance, with values approaching $1$ for an almost-perfect prediction.

\begin{table}
\caption{Results (mean and standard deviation for the coefficient of determination $R^2$) in the wind prediction task. Best result is highlighted in bold, second-best result in underlined.}
{\centering\hfill{}
	\setlength{\tabcolsep}{4pt}
	\renewcommand{\arraystretch}{1.5}
	\begin{small}
	\begin{tabular}{llc}   
	\toprule
	\textbf{Model} & & $\vect{R^2}$ \\ 
	\midrule
	\multirow{1}{*}{Linear}  & Linear     &                     $0.361 \pm    0.0227$ \\
	\midrule
	\multirow{2}{*}{Real-valued NNs} & 2R-NN (tanh)             &           $0.424   \pm   0.015$ \\
	& 2R-NN (ReLU)          &              $0.435   \pm   0.016$\\
	\midrule
	\multirow{4}{*}{CVNN} & C-NN (split-tanh)          &           $0.426  \pm    0.015$\\ 
	& C-NN (split-ReLU)       &              $0.438   \pm   0.016$\\
	& C-NN (AMP)               &             $0.431 \pm     0.014$\\
	& C-NN (CReLU)            &              $0.181    \pm  0.106$ \\
	& ModReLU-NN             &           $0.438  \pm    0.015$\\
	\midrule
	\multirow{3}{*}{Proposed CVNN} &  KAF-NN          &            $\vect{0.444 \pm     0.015}$\\
	& C-KAF-NN     &         $0.424  \pm    0.016$ \\
	& C-KAF-NN (Ind.) & $\underline{0.442   \pm   0.016}$\\
	\bottomrule
	\end{tabular}
	\end{small}
}
\hfill{}
\label{tab:results_wind}
\end{table}

Results for the experiment are reported in Table \ref{tab:results_wind}. We can see that, also in this scenario, the two best results are obtained by the proposed split-KAF and complex KAF neurons, significantly outperforming the other models.

\subsection{Experiment 3: complex-valued multi-class classification}
\label{sec:experiment_digits_classification}

We conclude our experimental evaluation by testing the proposed algorithms on a multi-class classification problem expressed in the complex domain. Following \cite{bouboulis2015complex}, we build the task by applying a two-dimensional fast Fourier transform (FFT) to the images in the well-known MNIST dataset,\footnote{\url{http://yann.lecun.com/exdb/mnist/}} comprising $60000$ $28 \times 28$ black-and-white images of handwritten digits split into ten classes. We then rank the coefficients of the FFT in terms of significance (by considering their mean absolute value), and keep only the $100$ most significant coefficients as input to the models. We compare a real-valued NN taking the real and the imaginary components of the coefficients as separate inputs, a CVNN with modReLU activation functions, and a CVNN employing the proposed split-KAF. All networks have a softmax activation function in their output layer. For the CVNNs, we use the following variation to handle the complex valued activations $\vect{h}$:
\begin{equation}
\text{softmax}_n(\vect{h}) = \frac{\exp\left\{\Re\left\{h_n\right\}^2+\Im\left\{h_n\right\}^2\right\}}{\sum_{t=1}^C \exp\left\{\Re\left\{h_t\right\}^2+\Im\left\{h_t\right\}^2\right\}} \,,
\end{equation}
where $\vect{h} \in \mathbb{C}^C$, and $C=10$ for our problem. All networks are then trained by minimizing the classical regularized cross-entropy formulation with the same optimizer as the last sections. We consider networks with three hidden layers having $100$ neurons each, whose regularization term is optimized via cross-validation separately. Results on the MNIST test set are provided in Table \ref{tab:results_mnist}.

\begin{figure}
\centering
\includegraphics[width=0.45\columnwidth,keepaspectratio]{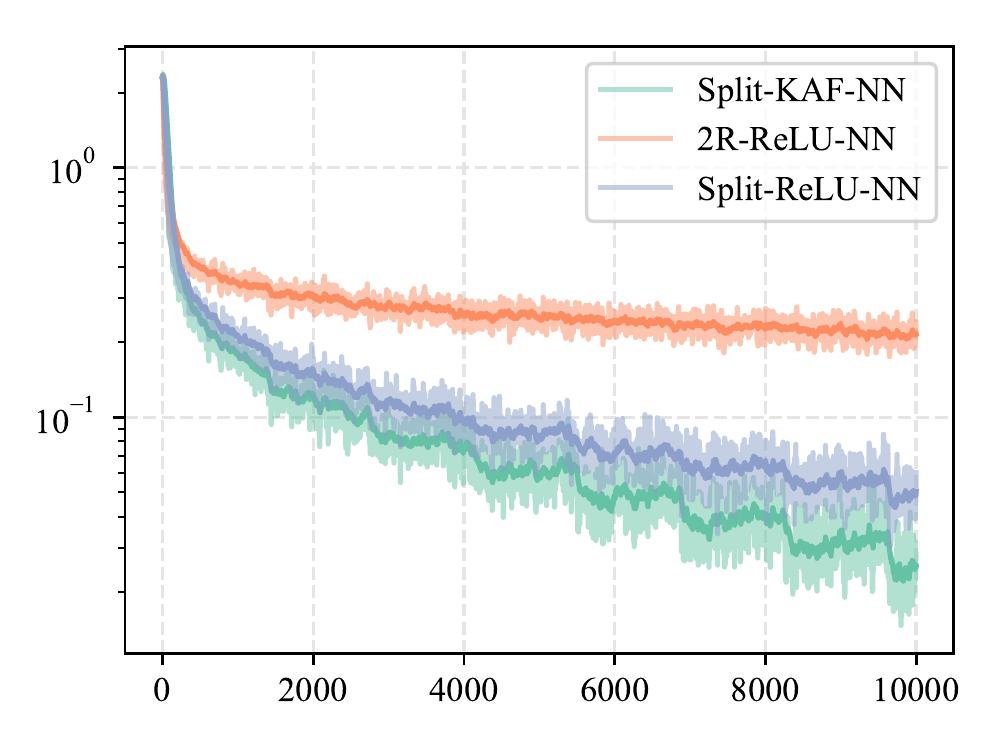}
\caption{Loss function evolution for the three algorithms on the complex-valued MNIST task (detail of the first 10000 iterations).}
\label{fig:mnist_convergence}
\end{figure}

We see that working in the complex domain results in significantly better performance when compared to working in the real domain. In addition, the proposed split-KAF can consistently obtain a better accuracy in this task than the ModReLU version. We show a representative evolution of the loss function in Fig. \ref{fig:mnist_convergence}, where we highlight the first 10000 iterations for readability.

\section{Conclusive remarks}
\label{sec:conclusions}

In this paper, we considered the problem of adapting the activation functions in a complex-valued neural network (CVNN). To this end, we proposed two different non-parametric models that extend the recently introduced kernel activation function (KAF) to the complex-valued case. The first model is a split configuration, where the real and the imaginary components of the activation are processed independently by two separate KAFs. In the second model, we directly redefine the KAF in the complex domain with the use of fully-complex kernels. We showed that CVNNs with adaptable functions can outperform neural networks with fixed functions in different benchmark problems including channel identification, wind prediction, and multi-class classification. For the fully-complex KAF, the independent kernel generally outperforms a naive complex Gaussian kernel without introducing significantly more complexity.

\begin{table}
\caption{Results (mean and standard deviation for the test accuracy) in the complex-valued MNIST task. Best result is highlighted in bold.}
{\centering\hfill{}
	\setlength{\tabcolsep}{4pt}
	\renewcommand{\arraystretch}{1.5}
	\begin{small}
	\begin{tabular}{lc}   
	\toprule
	\textbf{Model} & \textbf{Test accuracy [\%]} \\ 
	\midrule
	Real-valued NN & $92.39 \pm 0.10$ \\
	CVNN (ModReLU) & $95.92 \pm 0.18$ \\
	CVNN (Proposed split-KAF) & $\vect{97.21 \pm 0.34}$ \\
	\bottomrule
	\end{tabular}
	\end{small}
}
\hfill{}
\label{tab:results_mnist}
\end{table}

Several improvements over this framework are possible, most notably by leveraging over recent advances in the field of real-valued kernels (e.g. \cite{mansouri2017multiscale}) and complex-valued kernel regression and classification. One example is the use of pseudo-kernels \cite{boloix2017widely} to handle more efficiently the non-circularity in the signals propagated through the network. More in general, it would be interesting to extend other classes of non-parametric, real-valued activation functions (such as Maxout networks \cite{goodfellow2013maxout} or adaptive piecewise linear units \cite{agostinelli2014learning}) to the complex domain, or adapt the proposed complex KAFs to other types of NNs, such as convolutive architectures \cite{lecun2015deep,ren2017clustering}.

\section*{Acknowledgments}

The work of Simone Scardapane was supported in part by Italian MIUR, ``\textit{Progetti di Ricerca di Rilevante Interesse Nazionale}'',  GAUChO project, under Grant 2015YPXH4W\_004. The work of Steven Van Vaerenbergh was supported by the Ministerio de Econom{\'\i}a, Industria y Competitividad (MINECO) of Spain under grant TEC2014-57402-JIN (PRISMA). Amir Hussain was supported by the UK Engineering and Physical Science Research Council (EPSRC) grant no. EP/M026981/1.

\bibliographystyle{ieeetran}
\bibliography{biblio}

\begin{thebibliography}{10}
\providecommand{\url}[1]{#1}
\csname url@samestyle\endcsname
\providecommand{\newblock}{\relax}
\providecommand{\bibinfo}[2]{#2}
\providecommand{\BIBentrySTDinterwordspacing}{\spaceskip=0pt\relax}
\providecommand{\BIBentryALTinterwordstretchfactor}{4}
\providecommand{\BIBentryALTinterwordspacing}{\spaceskip=\fontdimen2\font plus
\BIBentryALTinterwordstretchfactor\fontdimen3\font minus
  \fontdimen4\font\relax}
\providecommand{\BIBforeignlanguage}[2]{{%
\expandafter\ifx\csname l@#1\endcsname\relax
\typeout{** WARNING: IEEEtran.bst: No hyphenation pattern has been}%
\typeout{** loaded for the language `#1'. Using the pattern for}%
\typeout{** the default language instead.}%
\else
\language=\csname l@#1\endcsname
\fi
#2}}
\providecommand{\BIBdecl}{\relax}
\BIBdecl

\bibitem{lecun2015deep}
Y.~LeCun, Y.~Bengio, and G.~Hinton, ``Deep learning,'' \emph{Nature}, vol. 521,
  no. 7553, pp. 436--444, 2015.

\bibitem{8264962}
N.~Shone, T.~N. Ngoc, V.~D. Phai, and Q.~Shi, ``A deep learning approach to
  network intrusion detection,'' \emph{IEEE Transactions on Emerging Topics in
  Computational Intelligence}, vol.~2, no.~1, pp. 41--50, Feb 2018.

\bibitem{zheng2017video}
K.~Zheng, W.~Q. Yan, and P.~Nand, ``Video dynamics detection using deep neural
  networks,'' \emph{IEEE Transactions on Emerging Topics in Computational
  Intelligence}, 2017.

\bibitem{hirose2003complex}
A.~Hirose, \emph{Complex-valued neural networks: theories and
  applications}.\hskip 1em plus 0.5em minus 0.4em\relax World Scientific, 2003,
  vol.~5.

\bibitem{schreier2010statistical}
P.~J. Schreier and L.~L. Scharf, \emph{Statistical signal processing of
  complex-valued data: the theory of improper and noncircular signals}.\hskip
  1em plus 0.5em minus 0.4em\relax Cambridge University Press, 2010.

\bibitem{mandic2007complex}
D.~P. Mandic, S.~Javidi, G.~Souretis, and V.~S. Goh, ``Why a complex valued
  solution for a real domain problem,'' in \emph{2007 IEEE Workshop on Machine
  Learning for Signal Processing}.\hskip 1em plus 0.5em minus 0.4em\relax IEEE,
  2007, pp. 384--389.

\bibitem{fisher1983complex}
B.~Fisher and N.~Bershad, ``The complex lms adaptive algorithm--transient
  weight mean and covariance with applications to the ale,'' \emph{IEEE
  Transactions on Acoustics, Speech, and Signal Processing}, vol.~31, no.~1,
  pp. 34--44, 1983.

\bibitem{bouboulis2011extension}
P.~Bouboulis and S.~Theodoridis, ``Extension of wirtinger's calculus to
  reproducing kernel hilbert spaces and the complex kernel lms,'' \emph{IEEE
  Transactions on Signal Processing}, vol.~59, no.~3, pp. 964--978, 2011.

\bibitem{tobar2012novel}
F.~A. Tobar, A.~Kuh, and D.~P. Mandic, ``A novel augmented complex valued
  kernel lms,'' in \emph{Sensor Array and Multichannel Signal Processing
  Workshop (SAM), 2012 IEEE 7th}.\hskip 1em plus 0.5em minus 0.4em\relax IEEE,
  2012, pp. 473--476.

\bibitem{boloix2017widely}
R.~Boloix-Tortosa, J.~J. Murillo-Fuentes, I.~Santos, and F.~P{\'e}rez-Cruz,
  ``Widely linear complex-valued kernel methods for regression,'' \emph{IEEE
  Transactions on Signal Processing}, vol.~65, no.~19, pp. 5240--5248, 2017.

\bibitem{scarpiniti2008generalized}
M.~Scarpiniti, D.~Vigliano, R.~Parisi, and A.~Uncini, ``Generalized splitting
  functions for blind separation of complex signals,'' \emph{Neurocomputing},
  vol.~71, no.~10, pp. 2245--2270, 2008.

\bibitem{georgiou1992complex}
G.~M. Georgiou and C.~Koutsougeras, ``Complex domain backpropagation,''
  \emph{IEEE Transactions on Circuits and Systems II: Analog and Digital Signal
  Processing}, vol.~39, no.~5, pp. 330--334, 1992.

\bibitem{kim2003approximation}
T.~Kim and T.~Adal{\i}, ``Approximation by fully complex multilayer
  perceptrons,'' \emph{Neural Computation}, vol.~15, no.~7, pp. 1641--1666,
  2003.

\bibitem{arjovsky2016unitary}
M.~Arjovsky, A.~Shah, and Y.~Bengio, ``Unitary evolution recurrent neural
  networks,'' in \emph{International Conference on Machine Learning}, 2016, pp.
  1120--1128.

\bibitem{danihelka2016associative}
I.~Danihelka, G.~Wayne, B.~Uria, N.~Kalchbrenner, and A.~Graves, ``Associative
  long short-term memory,'' \emph{arXiv preprint arXiv:1602.03032}, 2016.

\bibitem{guberman2016complex}
N.~Guberman, ``On complex valued convolutional neural networks,'' \emph{arXiv
  preprint arXiv:1602.09046}, 2016.

\bibitem{trabelsi2017deep}
C.~Trabelsi, O.~Bilaniuk, D.~Serdyuk, S.~Subramanian, J.~F. Santos, S.~Mehri,
  N.~Rostamzadeh, Y.~Bengio, and C.~J. Pal, ``Deep complex networks,''
  \emph{arXiv preprint arXiv:1705.09792}, 2017.

\bibitem{8109745}
A.~S. Shiva, M.~Gogate, N.~Howard, B.~Graham, and A.~Hussain, ``Complex-valued
  computational model of hippocampal ca3 recurrent collaterals,'' in \emph{2017
  IEEE 16th International Conference on Cognitive Informatics Cognitive
  Computing (ICCI*CC)}, July 2017, pp. 161--166.

\bibitem{leung1991complex}
H.~Leung and S.~Haykin, ``The complex backpropagation algorithm,'' \emph{IEEE
  Transactions on Signal Processing}, vol.~39, no.~9, pp. 2101--2104, 1991.

\bibitem{brandwood1983complex}
D.~Brandwood, ``A complex gradient operator and its application in adaptive
  array theory,'' in \emph{IEE Proceedings F - Communications, Radar and Signal
  Processing}, vol. 130, no.~1.\hskip 1em plus 0.5em minus 0.4em\relax IET,
  1983, pp. 11--16.

\bibitem{kreutz2009complex}
K.~Kreutz-Delgado, ``The complex gradient operator and the cr-calculus,''
  \emph{arXiv preprint arXiv:0906.4835}, 2009.

\bibitem{glorot2011deep}
X.~Glorot, A.~Bordes, and Y.~Bengio, ``Deep sparse rectifier neural networks.''
  in \emph{Aistats}, vol.~15, no. 106, 2011, p. 275.

\bibitem{maas2013rectifier}
A.~L. Maas, A.~Y. Hannun, and A.~Y. Ng, ``Rectifier nonlinearities improve
  neural network acoustic models,'' in \emph{Proc. 30th International
  Conference on Machine Learning (ICML)}, vol.~30, no.~1, 2013.

\bibitem{klambauer2017self}
G.~Klambauer, T.~Unterthiner, A.~Mayr, and S.~Hochreiter, ``Self-normalizing
  neural networks,'' \emph{arXiv preprint arXiv:1706.02515}, 2017.

\bibitem{ramachandran2017swish}
P.~Ramachandran, B.~Zoph, and Q.~V. Le, ``Swish: a self-gated activation
  function,'' \emph{arXiv preprint arXiv:1710.05941}, 2017.

\bibitem{nitta1997extension}
T.~Nitta, ``An extension of the back-propagation algorithm to complex
  numbers,'' \emph{Neural Networks}, vol.~10, no.~8, pp. 1391--1415, 1997.

\bibitem{he2015delving}
K.~He, X.~Zhang, S.~Ren, and J.~Sun, ``Delving deep into rectifiers: Surpassing
  human-level performance on imagenet classification,'' in \emph{Proc. IEEE
  International Conference on Computer Vision (ICCV)}, 2015, pp. 1026--1034.

\bibitem{jin2016deep}
X.~Jin, C.~Xu, J.~Feng, Y.~Wei, J.~Xiong, and S.~Yan, ``Deep learning with
  {S}-shaped rectified linear activation units,'' in \emph{Proc. Thirtieth AAAI
  Conference on Artificial Intelligence}, 2016.

\bibitem{goodfellow2013maxout}
I.~J. Goodfellow, D.~Warde-Farley, M.~Mirza, A.~Courville, and Y.~Bengio,
  ``Maxout networks,'' in \emph{Proc. 30th International Conference on Machine
  Learning (ICML)}, 2013.

\bibitem{agostinelli2014learning}
F.~Agostinelli, M.~Hoffman, P.~Sadowski, and P.~Baldi, ``Learning activation
  functions to improve deep neural networks,'' \emph{arXiv preprint
  arXiv:1412.6830}, 2014.

\bibitem{scardapane2016learning}
S.~Scardapane, M.~Scarpiniti, D.~Comminiello, and A.~Uncini, ``Learning
  activation functions from data using cubic spline interpolation,''
  \emph{arXiv preprint arXiv:1605.05509}, 2016.

\bibitem{scardapane2017kafnets}
S.~Scardapane, S.~Van~Vaerenbergh, S.~Totaro, and A.~Uncini, ``Kafnets:
  kernel-based non-parametric activation functions for neural networks,''
  \emph{arXiv preprint arXiv:1707.04035}, 2017.

\bibitem{bottou2016optimization}
L.~Bottou, F.~E. Curtis, and J.~Nocedal, ``Optimization methods for large-scale
  machine learning,'' \emph{arXiv preprint arXiv:1606.04838}, 2016.

\bibitem{xu2015convergence}
D.~Xu, H.~Zhang, and D.~P. Mandic, ``Convergence analysis of an augmented
  algorithm for fully complex-valued neural networks,'' \emph{Neural Networks},
  vol.~69, pp. 44--50, 2015.

\bibitem{zhang2016complex}
H.~Zhang and D.~P. Mandic, ``Is a complex-valued stepsize advantageous in
  complex-valued gradient learning algorithms?'' \emph{IEEE Transactions on
  Neural Networks and Learning Systems}, vol.~27, no.~12, pp. 2730--2735, 2016.

\bibitem{benvenuto1992complex}
N.~Benvenuto and F.~Piazza, ``On the complex backpropagation algorithm,''
  \emph{IEEE Transactions on Signal Processing}, vol.~40, no.~4, pp. 967--969,
  1992.

\bibitem{hirose1992continuous}
A.~Hirose, ``Continuous complex-valued back-propagation learning,''
  \emph{Electronics Letters}, vol.~28, no.~20, pp. 1854--1855, 1992.

\bibitem{virtue2017better}
P.~Virtue, S.~X. Yu, and M.~Lustig, ``Better than real: Complex-valued neural
  nets for mri fingerprinting,'' \emph{arXiv preprint arXiv:1707.00070}, 2017.

\bibitem{hofmann2008kernel}
T.~Hofmann, B.~Sch{\"o}lkopf, and A.~J. Smola, ``Kernel methods in machine
  learning,'' \emph{The Annals of Statistics}, pp. 1171--1220, 2008.

\bibitem{liu2011kernel}
W.~Liu, J.~C. Principe, and S.~Haykin, \emph{Kernel adaptive filtering: a
  comprehensive introduction}.\hskip 1em plus 0.5em minus 0.4em\relax John
  Wiley \& Sons, 2011, vol.~57.

\bibitem{aronszajn1950theory}
N.~Aronszajn, ``Theory of reproducing kernels,'' \emph{Transactions of the
  American Mathematical Society}, vol.~68, no.~3, pp. 337--404, 1950.

\bibitem{steinwart2006explicit}
I.~Steinwart, D.~Hush, and C.~Scovel, ``An explicit description of the
  reproducing kernel hilbert spaces of gaussian rbf kernels,'' \emph{IEEE
  Transactions on Information Theory}, vol.~52, no.~10, pp. 4635--4643, 2006.

\bibitem{maclaurin2015autograd}
D.~Maclaurin, D.~Duvenaud, and R.~P. Adams, ``Autograd: Effortless gradients in
  numpy,'' in \emph{ICML 2015 AutoML Workshop}, 2015.

\bibitem{bouboulis2015complex}
P.~Bouboulis, S.~Theodoridis, C.~Mavroforakis, and L.~Evaggelatou-Dalla,
  ``Complex support vector machines for regression and quaternary
  classification,'' \emph{IEEE Transactions on Neural Networks and Learning
  Systems}, vol.~26, no.~6, pp. 1260--1274, 2015.

\bibitem{goh2006complex}
S.~Goh, M.~Chen, D.~Popovi{\'c}, K.~Aihara, D.~Obradovic, and D.~Mandic,
  ``Complex-valued forecasting of wind profile,'' \emph{Renewable Energy},
  vol.~31, no.~11, pp. 1733--1750, 2006.

\bibitem{goh2004complex}
S.~L. Goh and D.~P. Mandic, ``A complex-valued rtrl algorithm for recurrent
  neural networks,'' \emph{Neural Computation}, vol.~16, no.~12, pp.
  2699--2713, 2004.

\bibitem{goh2005nonlinear}
------, ``Nonlinear adaptive prediction of complex-valued signals by
  complex-valued prnn,'' \emph{IEEE Transactions on Signal Processing},
  vol.~53, no.~5, pp. 1827--1836, 2005.

\bibitem{kuh2009applications}
A.~Kuh and D.~Mandic, ``Applications of complex augmented kernels to wind
  profile prediction,'' in \emph{Proc. 2009 IEEE International Conference on
  Acoustics, Speech and Signal Processing (ICASSP)}.\hskip 1em plus 0.5em minus
  0.4em\relax IEEE, 2009, pp. 3581--3584.

\bibitem{mansouri2017multiscale}
M.~Mansouri, M.~N. Nounou, and H.~N. Nounou, ``Multiscale kernel pls-based
  exponentially weighted-glrt and its application to fault detection,''
  \emph{IEEE Transactions on Emerging Topics in Computational Intelligence},
  2017.

\bibitem{ren2017clustering}
P.~Ren, W.~Sun, C.~Luo, and A.~Hussain, ``Clustering-oriented multiple
  convolutional neural networks for single image super-resolution,''
  \emph{Cognitive Computation}, pp. 1--14, 2017.

\end{thebibliography}

\end{document}